\documentclass[lettersize,journal]{IEEEtran}
\usepackage{amsmath,amsfonts}
\usepackage{algorithmic}
\usepackage{array}
\usepackage[caption=false,font=normalsize,labelfont=sf,textfont=sf]{subfig}
\usepackage{textcomp}
\usepackage{stfloats}
\usepackage{url}
\usepackage{verbatim}
\usepackage{graphicx}
\usepackage{balance}
\usepackage{booktabs}
\usepackage{multirow}
\usepackage{tikz}
\usetikzlibrary{shapes.geometric, arrows, positioning}
\usepackage[colorlinks=true, linkcolor=blue, citecolor=blue, urlcolor=blue]{hyperref}
\usepackage[center]{caption}
\usepackage[backend=biber, style=numeric, sorting=none]{biblatex}
\usepackage{times}                
\usepackage{booktabs}             
\usepackage{authblk}              
\def\BibTeX{{\rm B\kern-.05em{\sc i\kern-.025em b}\kern-.08em
    T\kern-.1667em\lower.7ex\hbox{E}\kern-.125emX}}

\title{\vspace{-1cm} \hrule height 4pt \vspace{0.7cm} 
\textbf{\huge Point Cloud to Mesh Reconstruction: Methods, Trade-offs, and Implementation Guide} 
\vspace{0.4cm} \hrule height 1pt}

\author[1]{\textbf{Fatima Zahra Iguenfer}}
\author[1]{\textbf{Achraf Hsain}}
\author[1]{\textbf{Hiba Amissa}}
\author[1]{\textbf{Yousra Chtouki}}

\affil[1]{School of Science and Engineering, Al Akhawayn University, Ifrane, Morocco}
\affil[ ]{\texttt{\{f.iguenfer, a.hsain, h.amissa, y.chtouki\}@aui.ma}}

\date{} 

\begin{document}

\maketitle

\begin{abstract}
Reconstructing meshes from point clouds is a fundamental task in computer vision with applications spanning robotics, autonomous systems, and medical imaging. Selecting an appropriate learning-based method requires understanding trade-offs between computational efficiency, geometric accuracy, and output constraints. This paper categorizes over fifteen methods into five paradigms---PointNet family, autoencoder architectures, deformation-based methods, point-move techniques, and primitive-based approaches---and provides practical guidance for method selection. We contribute: (1) a decision framework mapping input/output requirements to suitable paradigms, (2) a failure mode analysis to assist practitioners in debugging implementations, (3) standardized comparisons on ShapeNet benchmarks, and (4) a curated list of maintained codebases with implementation resources. By synthesizing both theoretical foundations and practical considerations, this work serves as an entry point for practitioners and researchers new to learning-based 3D mesh reconstruction.
\end{abstract}


\section{Introduction}

Point clouds represent three-dimensional spatial data created by sampling points from object surfaces. These digital collections offer adaptive storage and visualization capabilities across multiple levels of detail \cite{cao2018patch, li2015density}. Characterized by their flexibility, point clouds can be generated through various technologies, including 3D scanners, Light Detection and Ranging (LIDAR), structure-from-motion (SFM) techniques, and depth sensors such as Kinect and Xtion. Depending on the acquisition method, point clouds exhibit distinct characteristics in point density: SFM and photogrammetry-based approaches typically produce sparse point clouds, whereas 3D scanners, LIDAR, and depth sensors create more densely sampled representations \cite{liu2019deep}. This versatility, originating from the lack of topological and connectivity constraints characteristic of mesh-based models, renders point clouds particularly suitable for real-time applications \cite{camuffo2022recent}.

Three-dimensional meshes represent a prevalent discrete method for virtual surface and volume representation. Their inherent simplicity has led to widespread adoption, such that graphics processing units (GPUs) specialized in rendering images from 3D meshes are now integrated into personal computing devices including computers, tablets, and smartphones \cite{maglo20153d}. Triangular meshes offer an effective approach to 3D model representation, typically characterized by three fundamental data types: connectivity, geometry, and properties. Connectivity data define the adjacency relationships between vertices, providing the structural framework of the mesh. Geometry data precisely specify the spatial location of each vertex, while property data encompass additional attributes such as normal vectors, material reflectance, and texture coordinates \cite{peng2005technologies}.

Mesh reconstruction can therefore be defined as the transformation of point cloud data into 3D meshes. This technique finds applications in diverse domains, including 3D modeling and computer graphics, autonomous vehicles and robotics, architecture and urban planning, medical imaging, geospatial mapping, and augmented reality.

Although prior studies, such as \cite{Survey_Unstructured_data, reconstruction_tutorial, CGAL}, explore surface reconstruction from point clouds, they mainly focus on different representations, such as voxel grids, implicit surfaces, and volumetric representations. In addition, these studies often examine broader three-dimensional reconstruction methods or emphasize classical optimization-based approaches rather than learning-based techniques. Our work addresses this gap by specifically focusing on reconstructing mesh surfaces from point clouds using machine learning techniques. We organize these methods into five categories: PointNet Family, Autoencoder-based Architectures, Deformation-based Methods, Point-Move, and Primitive-based approaches.

\section{Method Selection Framework}

Before examining specific paradigms, practitioners must understand how to select an appropriate method based on their requirements. This section provides a structured decision framework to guide method selection.

\subsection{Input Characteristics}

The choice of reconstruction method depends on input data properties:

\textbf{Point Density:} Sparse point clouds (fewer than 1,000 points) benefit from completion-oriented methods like PCN \cite{yuan2018pcn}, while dense clouds (10,000+ points) can leverage direct triangulation approaches like PointTriNet \cite{sharp2020pointtrinet}.

\textbf{Noise Level:} Methods with robust encoders (PointNet-based architectures) handle moderate noise effectively. For high-noise scenarios, regression-based approaches \cite{ladicky2017} that aggregate local features provide better resilience.

\textbf{Completeness:} Partial point clouds require completion capabilities. The PCN family addresses this challenge, while direct triangulation methods assume relatively complete inputs.

\textbf{Single Object vs.\ Scene:} Most learning-based methods target single objects. Scene-level reconstruction typically requires segmentation preprocessing or specialized architectures.

\subsection{Output Requirements}

\textbf{Watertight Mesh Required:} If watertight (closed, manifold) meshes are essential, consider BSP-Net \cite{chen2020bspnet} which guarantees watertight outputs, or deformation-based methods starting from closed templates. AtlasNet \cite{groueix2018atlasnet} with sphere templates also produces closed surfaces.

\textbf{Sharp Feature Preservation:} For CAD-like models with sharp edges, primitive-based methods \cite{liu2023sharp} excel. Implicit methods and folding-based approaches tend to smooth sharp features.

\textbf{Real-time Constraints:} The Real-NVP deformation approach \cite{mansour2023} achieves 58Hz on 3,000-vertex meshes. Regression forests \cite{ladicky2017} offer millisecond-level reconstruction. Most deep learning methods require more substantial inference time.

\textbf{Topology Constraints:} Deformation-based methods preserve template topology. If arbitrary topology is needed, consider point-move or primitive assembly approaches.

\subsection{Decision Guidelines}

Based on common use cases, we recommend:

\begin{itemize}
    \item \textbf{General-purpose reconstruction:} AtlasNet provides a balance of quality and flexibility with multiple parameterizations.
    \item \textbf{Incomplete/partial data:} PCN family methods handle missing regions through learned completion.
    \item \textbf{CAD/mechanical parts:} Primitive-based methods preserve sharp geometric features better than implicit approaches.
    \item \textbf{Real-time robotics:} Deformation-based methods with template meshes offer fast inference when topology is known.
    \item \textbf{Direct triangulation:} PointTriNet when connectivity among existing points is needed without adding or removing vertices.
\end{itemize}

\section{The PointNet Family}

The PointNet-based algorithms represent an important advancement in 3D point cloud processing, providing scalable solutions for tasks such as 3D classification and segmentation. The original PointNet architecture \cite{qi2017pointnet}, introduced in 2017, directly processes point clouds using a symmetric function to ensure permutation invariance. This ability to handle raw point clouds without transformation into regular grids makes PointNet suitable for real-world 3D data processing, particularly in applications like autonomous driving, object detection, and robotics. Building on this foundation, PointNet++ \cite{qi2017pointnetpp}, introduced in the same year, extends PointNet by incorporating hierarchical feature learning to better capture local structures within point clouds. By processing the point cloud at multiple scales, PointNet++ improves performance on complex datasets where local geometric features are important. Subsequent developments, such as Generative PointNet \cite{xie2021generativepointnet}, introduce energy-based models for point cloud generation, reconstruction, and classification. Other works, like KNN-based Feature Learning Network \cite{luo2021knn}, explore semantic segmentation by leveraging K-nearest neighbors for feature learning, while PointNet++ for 3D Classification \cite{sheshappanavar2020} introduces improved local geometry capture through ellipsoid querying around centroids.

The PointNet family has established a foundation for advancements in point-cloud processing, which we explore further in subsequent sections. We begin with the original PointNet architecture, as it directly relates to mesh reconstruction from point clouds.

\subsection{PointNet}

The paper \cite{qi2017pointnet} presents a deep learning architecture designed to consume unordered point sets for tasks including object classification, part segmentation, and scene semantic parsing. Traditional deep learning approaches for 3D data typically transform point clouds into regular grid-based representations, such as 3D voxel grids, which introduce inefficiencies and quantization artifacts. PointNet operates directly on raw point cloud data, addressing the challenges of its unordered nature and ensuring permutation invariance through a symmetric function, specifically max pooling.

\begin{figure*}[ht]
\centering
\includegraphics[width=\textwidth]{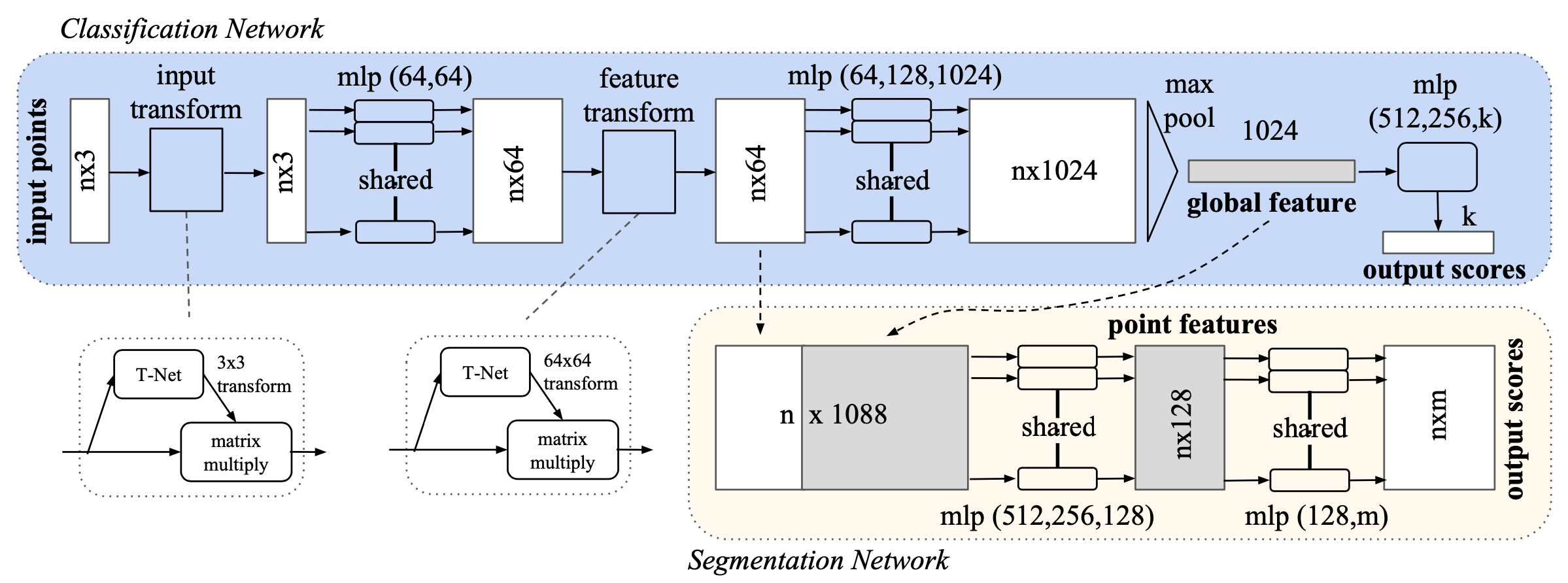}
\caption{PointNet Architecture \cite{qi2017pointnet}}
\label{fig:pointnet}
\end{figure*}

The PointNet architecture, presented in Figure~\ref{fig:pointnet}, consists of three core components:
\begin{itemize}
    \item A symmetric function to process unordered inputs.
    \item A combination of local and global information aggregation.
    \item A joint alignment network to align both input points and features for robust performance under geometric transformations.
\end{itemize}

The network produces a 1024-dimensional global feature vector that captures the essential geometric properties of the input point cloud. The network is computationally efficient, and experiments demonstrate that it outperforms traditional methods in segmentation, achieving state-of-the-art mean intersection over union (mIoU) \cite{rezatofighi2019giou} as shown in Table~\ref{tab:shapenet_results}.

\begin{table*}[h!]
\centering
\caption{Segmentation results on the ShapeNet \cite{chang2015shapenet} part dataset. Metric is mIoU (\%) on points. PointNet is compared to two traditional methods \cite{wu2014interactive, yi2016scalable} and a 3D fully convolutional network baseline proposed by \cite{qi2017pointnet}.}
\resizebox{0.95\textwidth}{!}{
\begin{tabular}{l|c|cccccccccccccccc}
\toprule
            & \textbf{mean} & \textbf{aero} & \textbf{bag} & \textbf{cap} & \textbf{car} & \textbf{chair} & \textbf{ear} & \textbf{guitar} & \textbf{knife} & \textbf{lamp} & \textbf{laptop} & \textbf{motor} & \textbf{mug} & \textbf{pistol} & \textbf{rocket} & \textbf{skate} & \textbf{table} \\
            &               &               &              &              &              &               & \textbf{phone} &                &                &                &                 &               &              &                &                & \textbf{board} &                \\
\midrule
\# shapes   & -             & 2690          & 76           & 55           & 898          & 3758          & 69            & 787            & 392            & 1547           & 451             & 202           & 184          & 283            & 66             & 152           & 5271           \\
\midrule
\cite{wu2014interactive}     & -             & 63.2          & -            & -            & -            & 73.5          & -             & -              & 85.4           & 74.4           & -               & -             & -            & -              & -              & -             & 74.8           \\
\cite{yi2016scalable}     & 81.4          & 81.0          & 78.4         & 77.7         & \textbf{75.7} & 87.6          & 61.9          & \textbf{92.0}  & 82.5           & \textbf{95.7}  & 70.6            & 91.9          & 85.9         & 53.1           & 69.8           & 75.3          & - \\
3DCNN       & 79.4          & 75.1          & 72.8         & 73.3         & 70.0         & 87.2          & 63.5          & 88.4           & 79.6           & 74.4           & 93.9            & 58.7          & 91.8         & 76.4           & 51.2           & 65.3          & 77.1           \\
PointNet        & \textbf{83.7} & \textbf{83.4} & \textbf{78.7} & \textbf{82.5} & 74.9         & \textbf{89.6} & \textbf{73.0} & 91.5           & \textbf{85.9}  & 80.8           & \textbf{95.3}   & \textbf{65.2} & \textbf{93.0} & \textbf{81.2}  & \textbf{57.9}  & \textbf{72.8} & \textbf{80.6}  \\
\bottomrule
\end{tabular}}
\label{tab:shapenet_results}
\end{table*}

The model's ability to learn both global and local features enables prediction of detailed per-point information, such as object part labels, based on both local geometry and global context. In addition to achieving strong results on benchmark datasets, PointNet demonstrates robustness to missing data and outliers, owing to its design that focuses on key points summarizing the shape of the input data \cite{qi2017pointnet}.

\subsection{PointTriNet: Learned Triangulation of 3D Point Sets}

The work \cite{sharp2020pointtrinet} introduces a differentiable and scalable method for generating triangulations among 3D points that integrates directly into learning pipelines. The approach iteratively applies two specialized neural networks: a classification network and a proposal network.

\begin{itemize}
    \item \textbf{Classification network:} Classifies query triangles using a PointNet-based network. For a given triangle, it gathers the 64 nearest points and triangles, encoding them relative to the query triangle, forming sets of nearby points and triangles processed by a multi-layer perceptron (MLP). The network outputs the probability of the query triangle belonging to the triangulation through localized, rigid-invariant encoding.
    \item \textbf{Proposal network:} Also a PointNet that identifies candidates by predicting the probability that a nearby vertex forms a neighboring triangle for an edge in a given triangle. Candidates are sampled using these probabilities, with initial scores derived from parent triangle probabilities.
\end{itemize}

Although PointTriNet produces meshes that may not always be watertight, when evaluated on sampled ShapeNet \cite{chang2015shapenet}, it outperformed classical methods such as the ball pivoting algorithm \cite{bernardini1999ball}, alpha-3, and alpha-5.

\subsection{PCN: Point Completion Network}

In \cite{yuan2018pcn}, the authors present a learning-based approach that operates directly on point clouds without voxelization or other structural assumptions, estimating complete geometry from sparse and incomplete point clouds. By avoiding voxelization, PCN achieves memory efficiency and prevents geometrical information loss that can occur with voxelization \cite{yuan2018pcn}. The PCN model follows an encoder-decoder architecture, where the encoder comprises multi-layer perceptrons (MLPs) and two stacked PointNet \cite{qi2017pointnet} layers, creating dense embeddings that are permutation-invariant and resistant to noise. Once a global feature vector is encoded, the decoder transforms this vector into both coarse and detailed point cloud representations \cite{yuan2018pcn}. The decoder leverages both fully connected layers \cite{achlioptas2018learning} and folding-based decoders \cite{yang2018foldingnet}, implemented in a multistage fashion for greater flexibility with fewer parameters.

The loss function comprises two components: the distance between the coarse output and ground truth, and a weighted distance between the detailed output and ground truth \cite{yuan2018pcn}. Both Chamfer distance and Earth Mover's distance \cite{fan2017point} are proposed as distance measures. For training, a synthetic dataset was created from CAD models in the ShapeNet dataset, focusing on eight categories: airplane, cabinet, car, chair, lamp, sofa, table, and vessel. A total of 16,384 points are uniformly sampled on each surface, and eight partial point clouds are generated per model by back-projecting into 2.5D space from random viewpoints. Out of 30,974 models selected from ShapeNet, 100 were used for validation and 150 for testing.

Results suggest that with an order of magnitude fewer parameters than alternatives \cite{yuan2018pcn}, PCN generalizes to unseen objects and real-world examples while being more scalable and robust than voxel-based methods \cite{yuan2018pcn}.

\begin{figure}[ht]
\centering
\includegraphics[scale=0.6]{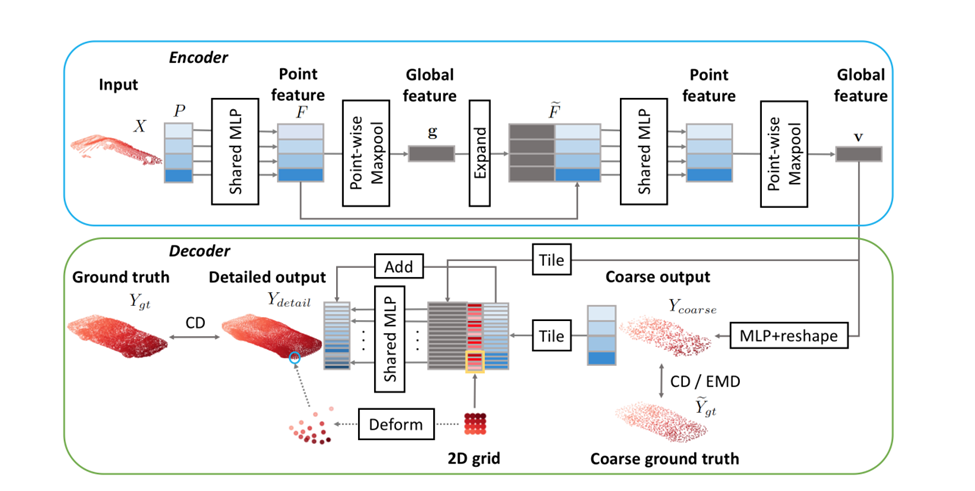}
\caption{PCN Architecture \cite{yuan2018pcn}}
\label{fig:pcn}
\end{figure}

\section{Autoencoder Architectures}

Another approach for point cloud to mesh reconstruction employs models based on the autoencoder architecture \cite{bank2021autoencoders}. These models consist of two main components:

\begin{itemize}
    \item \textbf{Encoder:} Processes the input point cloud, extracting important features and mapping the data to a lower-dimensional latent space. This latent representation captures the essential characteristics of the 3D object or scene.
    \item \textbf{Decoder:} Reconstructs the point cloud or mesh from the latent representation, producing a surface that represents the original shape. This reconstruction is guided by minimizing reconstruction error using loss functions such as Chamfer loss or Earth Mover's Distance (EMD) \cite{fan2017point}, which measure differences between generated points or meshes and targets.
\end{itemize}

In mesh reconstruction, autoencoders are particularly useful because they handle variations in point cloud density, noise, and partial data common in real-world applications. This section focuses on two approaches utilizing this architecture; later sections explore methods employing autoencoders within different paradigms.

\subsection{AtlasNet}

AtlasNet \cite{groueix2018atlasnet} generates 3D surfaces from feature representations, focusing on auto-encoding 3D shapes and reconstructing shapes from single-view images. The core idea is generating 3D surfaces using learnable parameterizations (charts), where each chart maps a 2D region $[0,1]^2$ to a 3D surface. These charts are learned through a Multilayer Perceptron (MLP) with ReLU activations \cite{agarap2019relu}, enabling generation of smooth and continuous surfaces directly from input point clouds or images. The method uses multiple charts to represent complex surfaces, combining them to cover the full shape. The model is trained using Chamfer loss, minimizing distance between generated points and target surface points.

For the auto-encoding task, the goal is to reconstruct a 3D surface from an input 3D point cloud. The model uses PointNet \cite{qi2017pointnet} as the encoder, transforming the input point cloud into a latent vector of dimension $k=1024$. The decoder consists of four fully connected layers (sizes: 1024, 512, 256, 128) with ReLU activations, except for the final layer using tanh. The decoder reconstructs a fixed-size point cloud of 2500 points during training, evenly sampled across learned parameterizations \cite{groueix2018atlasnet}. The model uses regular sampling during training to avoid overfitting and improve surface quality. This approach efficiently handles point clouds of varying sizes, processing up to 2500 points, although larger point clouds increase computational costs due to the quadratic nature of Chamfer loss.

During inference, AtlasNet generates high-resolution meshes by propagating patch-grid edges to 3D points. The simplest approach transfers a regular mesh from the unit square $[0,1]^2$ to the 3D surface, connecting points in 2D and mapping them to 3D coordinates. For typical use, 22,500 points are generated, though this can lead to non-closed meshes, small holes between parameterizations, and overlapping patches \cite{groueix2018atlasnet}. To address these drawbacks, the method generates a dense point cloud and applies Poisson Surface Reconstruction (PSR) \cite{kazhdan2013screened}. Alternatively, sampling points from a 3D sphere surface directly generates closed meshes without PSR, though quality depends on how well the surface can be represented by a sphere.

\subsection{3D Mesh Generation from a Defective Point Cloud using Style Transformation}

Another method \cite{tamata2022mesh} generates defect-free 3D meshes from defective point clouds by combining an autoencoder with an in-painting module and style transformation. The model uses PointNet as encoder and Neural Mesh Flow \cite{gupta2020neural} as decoder, both pretrained on point clouds without defects. The encoder captures features of defective point clouds, while the inpainting module transforms these features into defect-free ones using adversarial learning. The module incorporates noise to refine features and outputs a tensor via an MLP and max pooling layer.

Training occurs in two stages: learning geometrical structures using Chamfer distance loss to align the output mesh with the input point cloud, and training the inpainting module with GAN-based adversarial loss to convert defective styles to defect-free styles. This ensures effective reconstruction of defect-free meshes by complementing missing geometric structures.

\section{Deformation-based Methods}

We classify as deformation-based the methods that typically start with a template mesh, such as a sphere, and deform it to fit the desired shape. The deformation process adjusts mesh vertex positions but does not modify connectivity.

\subsection{Isomorphic Mesh Generation From Point Clouds With MLPs}

The paper \cite{miyauchi2022isomorphic} proposes a neural network architecture called isomorphic mesh generator (iMG) that reconstructs high-quality meshes from point clouds, even with noise or missing parts. Starting with a genus-zero (closed surface with no holes) spherical reference mesh (a subdivided icosahedron), iMG deforms the mesh to match the target shape using MLPs. The process involves three steps:

\begin{enumerate}
    \item \textbf{Global Mapping:} Uses an MLP to deform a sampled version of the reference mesh R(0) to fit the input point cloud, resulting in a coarse global approximation R(1) of the target object for further refinement.
    \item \textbf{Coarse Local Mapping:} Both reference mesh and point cloud are divided into 32 local regions. Each local mesh is independently refined and deformed to align with its corresponding local point cloud using MLPs. Refined local meshes are merged to improve overall accuracy.
    \item \textbf{Fine Local Mapping:} The process in step 2 is repeated with finer divisions for enhanced precision.
\end{enumerate}

Compared to AtlasNet, iMG demonstrates higher shape-recovery accuracy when generating isomorphic meshes using point clouds with and without noise. The network generates shapes close to ground truth even when input point clouds include missing parts obtained by mobile sensors.

\subsection{Meshing Point Clouds with Predicted Intrinsic-Extrinsic Ratio Guidance}

Another learning-based algorithm \cite{liu2020meshing} deforms candidate triangles to match point cloud inputs through several stages:

\textbf{Stage 1 -- Candidate Proposition:} Constructs a k-nearest neighbor (k-NN) graph on the input point cloud. Candidate triangle faces are proposed by considering combinations of each vertex and its k-nearest neighbors.

\textbf{Stage 2 -- Candidate Filtering:} Filters incorrect candidates using a deep neural network that predicts the Intrinsic-Extrinsic Ratio (IER)---the ratio of geodesic distance (intrinsic metric) to Euclidean distance (extrinsic metric) between two vertices. By applying a threshold on IER, invalid candidates are removed while valid triangles are retained.

\textbf{Stage 3 -- Sort and Merge:} Sorts remaining valid candidate triangles based on proximity to the surface and edge length, prioritizing those closer to the surface with smaller edge lengths. A greedy post-processing algorithm merges sorted triangles into the final mesh, ensuring no intersection between new and previously added faces.

\textbf{Stage 4 -- Extension to Reconstruction:} Extends the remeshing approach to full mesh reconstruction, where only the point cloud is given as input without a reference mesh.

The quantitative results show this method outperforms all baseline algorithms in F-score, Chamfer distance, and normal consistency across all categories \cite{liu2020meshing}. It significantly surpasses learning-based approaches like AtlasNet \cite{groueix2018atlasnet}, Deep Marching Cubes \cite{liao2018deep}, DeepSDF \cite{park2019deepsdf}, and Deep Geometric Prior \cite{williams2019deep}, which struggle with fine-grained details and generalization to unseen shapes.

\subsection{Fast Point Cloud to Mesh Reconstruction for Deformable Object Tracking}

In \cite{mansour2023}, the authors address a robotics challenge requiring real-time point cloud-to-mesh modeling of soft objects that deform upon contact with a robotic arm. They propose a model taking as input the point cloud representation of a deformed object along with a template mesh of the non-deformed object, generating the deformed mesh.

\begin{figure}[ht]
\centering
\includegraphics[scale=0.8]{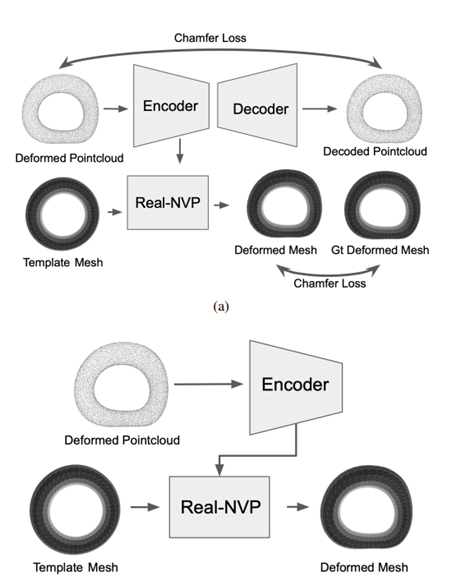}
\caption{Fast Point Cloud to Mesh Reconstruction for Deformable Object Tracking Pipeline \cite{mansour2023}}
\label{fig:deformable}
\end{figure}

The process begins by pretraining an autoencoder based primarily on convolutional neural network (CNN) layers to encode deformed point cloud information \cite{mansour2023}. The decoder reconstructs the original point cloud using Chamfer distance \cite{borgefors1984distance} reconstruction loss to minimize positional discrepancies \cite{mansour2023}. Once the encoder is pretrained, a conditional Real-NVP model \cite{dinh2016density} is employed. This model, consisting primarily of homeomorphic coupling blocks maintaining bijective mapping, takes the non-deformed template mesh along with encoder embeddings of the deformed point cloud. Real-NVP learns to deform the template mesh to match the deformed point cloud \cite{mansour2023}, chosen because it guarantees stable deformations without creating holes \cite{mansour2023}.

For training, the authors used synthetic data generated by applying random displacement fields to a 3$\times$3$\times$3 grid using Gaussian random variables, with interpolation using radial basis functions (RBF) \cite{park1991universal}. The model achieves a tracking rate of 58Hz on template meshes with 3,000 vertices and deformed point clouds of 5,000 points (0.017 seconds inference time on an Omen desktop machine) \cite{mansour2023}. The model generalizes to unseen deformations within object categories encountered during training \cite{mansour2023}. The training set includes six objects from the YCB benchmark dataset \cite{calli2015benchmarking}: scissors, hammer, foam brick, cleanser bottle, orange, and dice.

\subsection{3DN: 3D Deformation Network}

The paper \cite{wang20193dn} addresses insufficient 3D models in existing databases compared to abundant 2D image datasets. The authors propose an end-to-end network, 3D Deformation Network (3DN), designed to deform a source 3D mesh to match a target 3D model represented either as a 2D image or 3D point cloud \cite{wang20193dn}. This method estimates per-vertex displacement vectors for deformation, maintaining original mesh connectivity by altering only vertex positions \cite{wang20193dn}.

The key contribution of 3DN is deforming the source mesh without altering topology by predicting vertex displacement vectors directly in 3D space, achieving deformation while preserving mesh connectivity \cite{wang20193dn}. The architecture consists of three core components: a PointNet \cite{qi2017pointnet}-based encoder for extracting global features from the 3D source model, a target encoder (VGG \cite{simonyan2014vgg} for 2D images or PointNet for 3D point clouds) to generate global target features, and a decoder. The decoder takes concatenated global shape features from both source and target inputs, learning the mapping from original to deformed vertex locations through an MLP \cite{wang20193dn}. The authors also introduce a differentiable mesh sampling operator (DMSO) facilitating consistent point sampling from 3D meshes using barycentric coordinates \cite{wang20193dn}.

The training process incorporates multiple loss functions:
\begin{itemize}
    \item Chamfer Loss and Earth Mover's Distance (EMD) for overall shape similarity.
    \item Symmetry Loss: a combination of Chamfer and EMD losses ensuring symmetry preservation by comparing mirrored deformed output with target \cite{wang20193dn}.
    \item Mesh Laplacian Loss: preserves local geometric details by maintaining Laplacian coordinates of the source mesh.
    \item Local Permutation Invariant Loss: mitigates self-intersections, preserves local ordering of points, and enforces smooth deformations.
\end{itemize}

The network is trained using ShapeNet Core dataset \cite{chang2015shapenet} for 3D models and rendered views from \cite{choy20163d} for 2D images. The model is trained across ShapeNet Core's 13 shape categories \cite{wang20193dn}. Limitations include inability to accommodate topology changes required for some deformations, and computational expense of Chamfer and EMD metrics for high-resolution meshes \cite{wang20193dn}.

\subsection{FoldingNet: Point Cloud Auto-Encoder via Deep Grid Deformation}

The authors of \cite{yang2018foldingnet} present an end-to-end deep autoencoder for unsupervised learning on point clouds. The novelty lies in their folding-based decoder, introducing a unique approach to point cloud reconstruction. The folding operation is defined as ``the concatenation of replicated codewords to low-dimensional grid points, followed by a point-wise MLP'' \cite{yang2018foldingnet}.

The encoder utilizes a graph-based approach. The input is an $n \times 3$ matrix representing 3D spatial locations. For each point, the local covariance matrix is computed, vectorized into a $1 \times 9$ vector, and concatenated with 3D coordinates, forming an $n \times 12$ matrix. This matrix is processed through MLPs and graph-based max-pooling layers \cite{yang2018foldingnet}. The graph structure is a $k$-nearest neighbor graph ($k$-NNG) with $k = 16$. The encoder's final output is a $512$-dimensional codeword \cite{yang2018foldingnet}.

The decoder begins with a fixed $m \times 2$ grid of points, where $m = 2025$. This grid is concatenated with the codeword replicated $m$ times, forming an $m \times 514$ matrix. The decoder performs two sequential folding operations using 3-layer MLPs, mapping the grid from 2D to 3D space \cite{yang2018foldingnet}. This folding-based decoder achieves significant parameter efficiency, using approximately 7\% of parameters required by fully connected decoders \cite{achlioptas2018learning}. The authors state that a 2-layer perceptron can construct arbitrary point clouds from a 2D grid using folding \cite{yang2018foldingnet}. The Chamfer distance is computed bi-directionally as reconstruction loss.

Their autoencoder achieved Linear SVM classification accuracy of $88.4\%$ on ModelNet40 and $94.4\%$ on ModelNet10 \cite{yang2018foldingnet}, with preprocessing similar to \cite{socher2012convolutional}, normalizing point clouds to a unit sphere containing 2048 points per model. This surpassed other unsupervised methods on major datasets.

\subsection{CaDeX}

In \cite{lei2022cadex}, the authors address limitations in deformable surface representation by introducing CaDeX (Canonical Deformation Coordinate Space). CaDeX utilizes novel factorization of deformation via continuous bijective canonical maps (homeomorphisms) between frames, anchored to a learned canonical shape. CaDeX reconstructs sequences of surfaces from point cloud observations of deforming instances while enabling interframe correspondence through the canonical shape. A canonical map links each deformed frame to this shared shape, with CaDeX representing global 3D coordinates consistent across time \cite{lei2022cadex}. The maps are invertible, implemented using conditional Real-NVP \cite{dinh2016density} or NICE \cite{dinh2014nice} frameworks for efficient homeomorphism learning.

Each deformed frame has a unique canonical map conditioned on a deformation embedding derived via PointNet or ST-PointNet \cite{tang2021learning} encoders. CaDeX ensures key properties including cycle consistency, topology preservation, volume conservation with NICE, and continuity if the deformation embedding is time-continuous \cite{lei2022cadex}. The canonical shape is modeled using an occupancy network \cite{mescheder2019occupancy}, with geometry embeddings aggregated across frames. During training, a reconstruction loss $L_R$ ensures accurate surface modeling, with optional correspondence loss $L_C$ when supervision is available.

Contributions include a novel general representation and architecture for dynamic surfaces jointly solving canonical shape and consistent deformation problems, learnable continuous bijective canonical maps ensuring cycle consistency and topology preservation, and a novel solution to dynamic surface reconstruction and correspondence tasks from sparse point clouds or depth views. The method achieves state-of-the-art performance in modeling various deformable categories \cite{lei2022cadex}. Limitations include trembling in output when input undergoes large discontinuities and occasional topology changes in real-world deformations \cite{lei2022cadex}.

\section{Point Move Methods}

\subsection{From Point Clouds to Mesh using Regression}

\begin{figure*}[ht]
\centering
\includegraphics[scale=1]{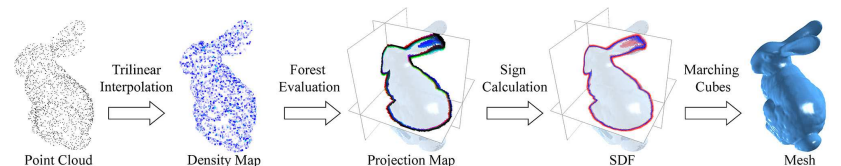}
\caption{Point Cloud to Mesh using Regression Pipeline \cite{ladicky2017}}
\label{fig:regression}
\end{figure*}

In \cite{ladicky2017}, the authors employ a regression forest model learning a function mapping feature vectors to corresponding projection difference vectors---vectors from each grid point to the closest surface point. Rather than using density at a single grid point, which would insufficiently capture local density trends, the authors use total densities: aggregation of densities from multiple randomly generated 3D boxes across various regions around each point \cite{ladicky2017}. This aggregation helps determine local density trends more robustly. The feature vector for grid point $x$ is an aggregation of regional densities across multiple randomly generated 3D boxes in the neighborhood of $x$.

Using these feature vectors, the regression forest \cite{breiman2001random} learns to approximate projection difference vectors for each grid point, enabling prediction of each point's relative position to the surface without requiring surface normals \cite{ladicky2017}. Training minimizes a squared error loss function. Each tree is trained by continuously splitting data to reduce variance, and the final prediction is obtained by averaging results across all trees \cite{ladicky2017}.

To enhance computational efficiency, the authors implemented an octree-based evaluation strategy. This reduces unnecessary calculations by first splitting space into $8 \times 8 \times 8$ cells, then testing subdivisions for surface point presence and only refining evaluation where needed. By extending each subdivision by $K$ cells in each direction ($K = 8$), the authors avoided approximately 90\% of forest evaluations \cite{ladicky2017}.

The authors also considered two alternative methods: (1) Direct regression of absolute distance, which had drawbacks including requiring iso-surface extraction at a non-zero slack value and producing two surfaces from unsigned distance; (2) Regression of signed distance, which would require normals in the feature vector, significantly decreasing computational efficiency.

The final pipeline: (1) Trilinear interpolation to obtain point cloud density map, (2) Forest evaluation for density map, (3) Sign calculation to obtain signed distance field (SDF), (4) Marching cubes algorithm for final mesh. The authors used synthetic data from ModelNet40 dataset \cite{wu2015modelnet} (Flowerpot, Lamp, Plant, Sink, Toilet, and Vase models), with testing on real-world data captured using a mobile phone \cite{ladicky2017}. Advantages include reconstructing high-resolution meshes in milliseconds, quality comparable to state of the art, and applicability to arbitrarily large point clouds.

\subsection{3D Reconstruction of Point Cloud Based on Point-Move}

The paper \cite{yi2024pointmove} presents an unsupervised 3D reconstruction method called Point-Move, reconstructing 3D models from point clouds without requiring 3D ground truth labels. The method's core is a deep learning network learning the Signed Distance Function (SDF) from raw point cloud data. It leverages a PointNet-based architecture to extract local features, then iteratively refines point positions, moving them toward the underlying surface to generate detailed and accurate meshes.

Experiments on ShapeNet and ModelNet show Point-Move outperforms existing point cloud reconstruction techniques in both surface accuracy and ability to handle complex geometries. The method demonstrates strong performance even when labeled 3D data is scarce, making it suitable for applications where annotated data is limited.

\section{Primitive Based Methods}

\subsection{Sharp Feature-Preserving 3D Mesh Reconstructions}

In \cite{liu2023sharp}, the authors address the challenge of accurately capturing sharp, continuous edges in 3D mesh reconstruction from point clouds, proposing a novel framework enhancing sharp feature preservation. This framework incorporates an advanced deep learning-based primitive detection module alongside innovative mesh fitting, splitting, and selection modules. The framework takes point clouds as input and outputs high-fidelity, watertight mesh models \cite{liu2023sharp}.

The primary component is the primitive detection module. The authors utilize HPNet \cite{yan2021hpnet} as a base detector with significant enhancements: replacing DGCNN \cite{wang2019dynamic} with PointNeXt-b \cite{qian2022pointnext} to address throughput limitations, switching from Adam optimizer to AdamW, and integrating cosine learning rate decay with label smoothing. These changes increase segmentation mean Intersection-over-Union (IoU) and primitive type mean IoU from $85.24/91.04$ to $88.42/92.85$ on the ABCParts benchmark \cite{sharma2020parsenet}, with three-fold throughput improvement \cite{liu2023sharp}. This module processes input point clouds to generate K primitive segments, applying refined clustering based on normal angles to counteract over-segmentation, discarding patches below minimum size $N_{\min}$, and computing convex hulls for remaining patches \cite{liu2023sharp}.

The framework's second stage, mesh fitting and splitting, comprises four steps: mesh fitting (using the method from \cite{huang2019variational}), intersection line detection, pairwise splitting, and partitioning triangles in non-intersecting areas. Intersection detection computes axis-aligned bounding boxes \cite{kettner2021intersecting} for each triangle and uses collision detection to identify intersecting pairs. The final selection module frames optimal subset selection as choosing desired meshes from candidates \cite{liu2023sharp}.

\begin{figure}[ht]
\centering
\includegraphics[scale=0.5]{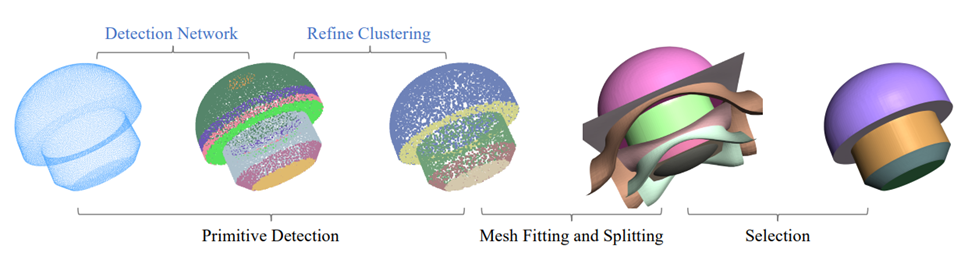}
\caption{Sharp Feature-Preserving 3D Mesh Reconstruction Pipeline \cite{liu2023sharp}}
\label{fig:sharp}
\end{figure}

The authors tested their framework using ABCParts \cite{sharma2020parsenet} and Thingi10K \cite{zhou2016thingi10k} datasets, evaluating with three metrics: Seg-IoU (similarity between predicted patches and ground truth), Type-IoU (classification accuracy of predicted primitive types), and Throughput (instances per second). The framework achieves Seg-IoU of $88.42\%$, Type-IoU of $92.85\%$, and Throughput of 28 instances/sec, with robustness against noise and incomplete data \cite{liu2023sharp}. Future improvements depend on expanding available 3D datasets beyond CAD models.

\subsection{BSP-Net: Generating Compact Meshes via Binary Space Partitioning}

BSP-Net \cite{chen2020bspnet} introduces a framework for generating compact, watertight 3D polygonal meshes directly from input data such as point clouds or voxel grids. Addressing limitations of methods relying on implicit functions and computationally expensive iso-surfacing routines, the authors leverage Binary Space Partitioning (BSP) to represent 3D shapes as collections of convex components. BSP-Net's architecture comprises three stages: predicting hyperplanes through an MLP, grouping planes into convex shapes using binary space partitioning tree structure, and combining convex parts to form complete meshes. This unsupervised learning approach eliminates need for ground-truth convex decompositions and enables generation of meshes that are compact, computationally efficient, and capable of capturing sharp geometric features. BSP-Net outperforms state-of-the-art methods in auto-encoding and single-view reconstruction by achieving better trade-off between mesh fidelity and complexity. The model is particularly suited for applications requiring low-polygon, high-detail 3D representations.

\section{Comparative Analysis}

This section provides structured comparisons of surveyed methods to guide practitioners in selecting appropriate approaches for their requirements.

\subsection{Paradigm Comparison}

Table~\ref{tab:paradigm_comparison} summarizes the methodological characteristics, loss functions, datasets, and key results across paradigms.

\begin{table*}[ht]
\centering
\caption{Comparison of Mesh Reconstruction Approaches by Paradigm}
\resizebox{\textwidth}{!}{
\begin{tabular}{@{}p{2.5cm}p{4.5cm}p{3cm}p{3cm}p{4cm}@{}}
\toprule
\textbf{Approach} & \textbf{Methodology} & \textbf{Loss Functions} & \textbf{Datasets} & \textbf{Key Results} \\
\midrule
PointNet Family & Processes point clouds directly using symmetric functions, hierarchical features, triangulation, and shape completion with energy-based models & Chamfer distance, Earth Mover's Distance & ShapeNet, ModelNet40 & Shape completion and improved geometry learning; PointNet achieves 83.7\% mean mIoU on ShapeNet segmentation \\
\midrule
Autoencoder & Encoder-decoder with latent space representation; examples include AtlasNet and FoldingNet & Chamfer Loss, Earth Mover's Distance & ShapeNet, ModelNet & AtlasNet reconstructed 3D meshes with demonstrated generalization to unseen categories \\
\midrule
Deformation-Based & Iteratively deforms template mesh to fit point cloud; examples include 3DN, Real-NVP, iMG & Chamfer Loss, Mesh Laplacian Loss & ShapeNet Core, YCB Object Set & 3DN maintained topology and achieved detailed deformations; Real-NVP processed at 58Hz for 3000-vertex meshes; FoldingNet: 88.4\% accuracy on ModelNet40 \\
\midrule
Point-Move & Iteratively refines point positions to reconstruct mesh from Signed Distance Function (SDF) & Chamfer Loss, Squared Error & ShapeNet, ModelNet & Regression forests achieve millisecond-level reconstruction; Point-Move handles complex geometries \\
\midrule
Primitive-Based & Detecting and fitting geometric primitives; uses segmentation and clustering techniques & Segmentation IoU, Chamfer Loss & ABCParts, Thingi10K, ModelNet & Sharp Features: 88.42\% Seg-IoU, 92.85\% Type-IoU, 28 instances/sec \\
\bottomrule
\end{tabular}}
\label{tab:paradigm_comparison}
\end{table*}

\subsection{Method Characteristics Comparison}

Table~\ref{tab:method_comparison} presents a practical comparison of specific methods across important dimensions.

\begin{table*}[ht]
\centering
\caption{Practical Comparison of Mesh Reconstruction Methods}
\resizebox{\textwidth}{!}{
\begin{tabular}{@{}lccccccc@{}}
\toprule
\textbf{Method} & \textbf{Paradigm} & \textbf{Watertight} & \textbf{Sharp Features} & \textbf{Handles Noise} & \textbf{Real-time} & \textbf{Framework} & \textbf{Code Available} \\
\midrule
PointNet \cite{qi2017pointnet} & Encoder & N/A & N/A & Yes & No & TensorFlow & \href{https://github.com/charlesq34/pointnet}{github.com/charlesq34/pointnet} \\
PointNet++ \cite{qi2017pointnetpp} & Encoder & N/A & N/A & Yes & No & TensorFlow & \href{https://github.com/charlesq34/pointnet2}{github.com/charlesq34/pointnet2} \\
PointTriNet \cite{sharp2020pointtrinet} & PointNet Family & No & Moderate & Yes & No & PyTorch & \href{https://github.com/nmwsharp/learned-triangulation}{github.com/nmwsharp/learned-triangulation} \\
PCN \cite{yuan2018pcn} & PointNet Family & No & No & Yes & No & TensorFlow & \href{https://github.com/wentaoyuan/pcn}{github.com/wentaoyuan/pcn} \\
AtlasNet \cite{groueix2018atlasnet} & Autoencoder & Optional$^*$ & No & Moderate & No & PyTorch & \href{https://github.com/ThibaultGROUEIX/AtlasNet}{github.com/ThibaultGROUEIX/AtlasNet} \\
FoldingNet \cite{yang2018foldingnet} & Autoencoder & No & No & Moderate & No & Caffe/PyTorch & \href{https://github.com/merlresearch/FoldingNet}{github.com/merlresearch/FoldingNet} \\
3DN \cite{wang20193dn} & Deformation & Yes$^\dagger$ & Moderate & Yes & No & PyTorch & --- \\
Real-NVP \cite{mansour2023} & Deformation & Yes$^\dagger$ & No & Yes & Yes (58Hz) & --- & --- \\
iMG \cite{miyauchi2022isomorphic} & Deformation & Yes$^\dagger$ & Moderate & Yes & No & --- & --- \\
Regression Forest \cite{ladicky2017} & Point-Move & No & No & Yes & Yes (ms) & --- & --- \\
Point-Move \cite{yi2024pointmove} & Point-Move & No & No & Yes & No & --- & --- \\
Sharp Features \cite{liu2023sharp} & Primitive & Yes & Yes & Yes & No & --- & --- \\
BSP-Net \cite{chen2020bspnet} & Primitive & Yes & Yes & Moderate & No & TF/PyTorch & \href{https://github.com/czq142857/BSP-NET-original}{github.com/czq142857/BSP-NET-original} \\
\bottomrule
\end{tabular}}
\begin{flushleft}
\footnotesize{$^*$Watertight when using sphere template; $^\dagger$Preserves template topology}
\end{flushleft}
\label{tab:method_comparison}
\end{table*}

\subsection{When Each Paradigm Excels}

\textbf{PointNet Family:} Best suited for scenarios requiring direct point cloud processing with permutation invariance. PCN excels at shape completion from partial data, while PointTriNet provides direct triangulation without intermediate representations. These methods generalize well across object categories through local geometric learning.

\textbf{Autoencoder Architectures:} AtlasNet and FoldingNet learn compact shape representations suitable for shape analysis and generation. AtlasNet's multiple chart approach handles complex topologies, while FoldingNet's parameter efficiency (approximately 7\% of fully-connected alternatives) is attractive for resource-constrained settings. Both achieve strong transfer classification performance.

\textbf{Deformation-based Methods:} Optimal when template topology is known or can be reasonably approximated. They maintain mesh connectivity throughout deformation, avoiding self-intersections common in other approaches. The 58Hz tracking rate achieved by Real-NVP demonstrates practical viability for interactive systems.

\textbf{Point-Move Methods:} The regression forest approach \cite{ladicky2017} achieves millisecond-level reconstruction suitable for interactive applications. These methods work well with arbitrarily large point clouds through octree-based evaluation strategies.

\textbf{Primitive-based Methods:} BSP-Net and sharp feature methods excel on CAD-like models. BSP-Net guarantees watertight, low-polygon outputs with sharp edges. The sharp feature method \cite{liu2023sharp} specifically targets mechanical parts where edge preservation is critical.

Understanding common failure patterns helps practitioners debug implementations and select appropriate methods. This section documents failure modes observed across paradigms.

\subsection{PointNet Family Methods}

\begin{tabular}{@{}p{0.18\linewidth} p{0.78\linewidth}@{}}
\textbf{Failure:} & Loss of fine geometric details in global feature aggregation. \\
\textbf{Symptom:} & Reconstructions appear overly smooth, missing thin structures like chair legs or airplane wings. \\
\textbf{Mitigation:} & Use hierarchical variants (PointNet++) that capture multi-scale features, or increase the number of critical points in the max-pooling operation. \\
\end{tabular}

\vspace{6pt}

\begin{tabular}{@{}p{0.18\linewidth} p{0.78\linewidth}@{}}
\textbf{Failure:} & Poor performance on out-of-distribution point densities. \\
\textbf{Symptom:} & Model trained on 2048-point clouds fails on sparser or denser inputs. \\
\textbf{Mitigation:} & Augment training data with varying point densities; use density-invariant sampling during inference. \\
\end{tabular}

\subsection{Autoencoder Methods}

\begin{tabular}{@{}p{0.18\linewidth} p{0.78\linewidth}@{}}
\textbf{Failure:} & Blurry reconstructions and missing thin structures. \\
\textbf{Symptom:} & Chamfer loss plateaus but visual quality remains poor; surfaces appear over-smoothed. \\
\textbf{Mitigation:} & Increase latent dimension; add Earth Mover's Distance loss; use more parameterization patches in AtlasNet. \\
\end{tabular}

\vspace{6pt}

\begin{tabular}{@{}p{0.18\linewidth} p{0.78\linewidth}@{}}
\textbf{Failure:} & Holes between AtlasNet patches. \\
\textbf{Symptom:} & Visible gaps at boundaries between learned parameterizations. \\
\textbf{Mitigation:} & Apply Poisson Surface Reconstruction post-processing; use sphere template for closed surfaces; increase sampling density at patch boundaries. \\
\end{tabular}

\subsection{Deformation-based Methods}

\begin{tabular}{@{}p{0.18\linewidth} p{0.78\linewidth}@{}}
\textbf{Failure:} & Self-intersections when source/target topology differs significantly. \\
\textbf{Symptom:} & Folded mesh regions, inverted normals, non-manifold geometry. \\
\textbf{Mitigation:} & Use Laplacian regularization; check genus compatibility between template and target; apply Local Permutation Invariant Loss (as in 3DN). \\
\end{tabular}

\vspace{6pt}

\begin{tabular}{@{}p{0.18\linewidth} p{0.78\linewidth}@{}}
\textbf{Failure:} & Template topology limitations. \\
\textbf{Symptom:} & Cannot represent shapes with different genus (e.g., sphere template cannot form a torus). \\
\textbf{Mitigation:} & Select appropriate template topology; consider methods that can modify connectivity. \\
\end{tabular}

\vspace{6pt}

\begin{tabular}{@{}p{0.18\linewidth} p{0.78\linewidth}@{}}
\textbf{Failure:} & CaDeX trembling artifacts. \\
\textbf{Symptom:} & Output exhibits trembling when input undergoes large discontinuities. \\
\textbf{Mitigation:} & Ensure smooth deformation embeddings; consider temporal smoothing for sequence inputs. \\
\end{tabular}

\subsection{Primitive-based Methods}

\begin{tabular}{@{}p{0.18\linewidth} p{0.78\linewidth}@{}}
\textbf{Failure:} & Over-segmentation on organic shapes. \\
\textbf{Symptom:} & Excessive primitive count, discontinuous surfaces, unnatural appearance on smooth objects. \\
\textbf{Mitigation:} & Adjust clustering thresholds; use these methods primarily for CAD/mechanical models; apply smoothing post-processing for organic shapes. \\
\end{tabular}

\vspace{6pt}

\begin{tabular}{@{}p{0.18\linewidth} p{0.78\linewidth}@{}}
\textbf{Failure:} & BSP-Net convergence issues. \\
\textbf{Symptom:} & Training loss oscillates; output convexes don't form coherent shapes. \\
\textbf{Mitigation:} & Use progressive training phases (as recommended by authors); adjust the number of convex components; verify input normalization. \\
\end{tabular}

\section{Implementation Resources}
\label{sec:resources}

This section provides practitioners with curated resources for implementing the surveyed methods.

\subsection{Official Code Repositories}

Table~\ref{tab:code_repos} lists maintained codebases with their frameworks and status.

\begin{table*}[ht]
\centering
\caption{Code Repositories for Surveyed Methods}
\begin{tabular}{@{}lll@{}}
\toprule
\textbf{Method} & \textbf{Framework} & \textbf{Repository} \\
\midrule
PointNet & TensorFlow 1.x & \url{https://github.com/charlesq34/pointnet} \\
PointNet++ & TensorFlow 1.x & \url{https://github.com/charlesq34/pointnet2} \\
PCN & TensorFlow 1.12 & \url{https://github.com/wentaoyuan/pcn} \\
AtlasNet & PyTorch & \url{https://github.com/ThibaultGROUEIX/AtlasNet} \\
PointTriNet & PyTorch & \url{https://github.com/nmwsharp/learned-triangulation} \\
FoldingNet & Caffe/PyTorch & \url{https://github.com/merlresearch/FoldingNet} \\
BSP-Net (TF1) & TensorFlow 1.15 & \url{https://github.com/czq142857/BSP-NET-original} \\
BSP-Net (TF2) & TensorFlow 2.0 & \url{https://github.com/czq142857/BSP-NET-tf2} \\
BSP-Net (PyTorch) & PyTorch 1.2+ & \url{https://github.com/czq142857/BSP-NET-pytorch} \\
\bottomrule
\end{tabular}
\label{tab:code_repos}
\end{table*}

\subsection{Benchmark Datasets}

Table~\ref{tab:datasets} summarizes commonly used datasets for training and evaluation.

\begin{table}[ht]
\centering
\caption{Benchmark Datasets for 3D Reconstruction}
\resizebox{\columnwidth}{!}{\begin{tabular}{@{}lccl@{}}
\toprule
\textbf{Dataset} & \textbf{Models} & \textbf{Categories} & \textbf{URL} \\
\midrule
ShapeNetCore v1 & 51,300 & 55 & \href{https://shapenet.org}{shapenet.org} \\
ShapeNetCore v2 & 51,300 & 57 & \href{https://shapenet.org}{shapenet.org} \\
ModelNet40 & 12,311 & 40 & \href{https://modelnet.cs.princeton.edu}{modelnet.cs.princeton.edu} \\
ModelNet10 & 4,899 & 10 & \href{https://modelnet.cs.princeton.edu}{modelnet.cs.princeton.edu} \\
ABCParts & 1M+ CAD & Primitives & \href{https://deep-geometry.github.io/abc-dataset}{deep-geometry.github.io/abc-dataset} \\
Thingi10K & 10,000 & Various & \href{https://ten-thousand-models.appspot.com}{ten-thousand-models.appspot.com} \\
\bottomrule
\end{tabular}}
\label{tab:datasets}
\end{table}

\subsection{Evaluation Tools and Libraries}

\textbf{PyTorch3D} (\href{https://pytorch3d.org}{pytorch3d.org}): Provides differentiable Chamfer distance, mesh operations, and rendering. Recommended for PyTorch-based pipelines.

\textbf{Open3D} (\href{https://www.open3d.org}{open3d.org}): Comprehensive point cloud processing, visualization, and classical reconstruction algorithms. Useful for data preprocessing and baseline comparisons.

\textbf{Trimesh} (\href{https://trimesh.org}{trimesh.org}): Mesh I/O, watertight checking, boolean operations. Essential for validating mesh quality and post-processing.

\textbf{Kaolin} (\href{https://github.com/NVIDIAGameWorks/kaolin}{github.com/NVIDIAGameWorks/kaolin}): NVIDIA's library for 3D deep learning with ShapeNet dataloaders and differentiable rendering.

\subsection{Computational Considerations}

Based on reported experimental setups from original papers:

\textbf{Training Time:} Most methods require 24--72 hours on a single GPU (V100/A100 class) for full ShapeNet training. PCN and AtlasNet train faster due to efficient architectures.

\textbf{Inference Time:} Varies significantly---Real-NVP achieves 17ms per object; most deep learning methods require 100ms--1s per object; regression forests achieve sub-millisecond times.

\textbf{Memory Requirements:} Chamfer distance computation is quadratic in point count; methods using 2500+ points require 8--16GB GPU memory during training. BSP-Net's convex operations are memory-efficient.

\section{Conclusion}

This paper examined learning-based approaches for mesh reconstruction from point clouds, organizing them into five paradigms: PointNet family, autoencoder architectures, deformation-based methods, point-move techniques, and primitive-based approaches. Each paradigm offers distinct trade-offs that practitioners must consider when selecting methods for specific applications.

\textbf{PointNet-family methods} provide robust, permutation-invariant processing of raw point clouds. PCN's coarse-to-fine completion strategy handles partial inputs effectively, while PointTriNet offers differentiable triangulation that integrates directly into learning pipelines. These methods generalize well across object categories through local geometric learning.

\textbf{Autoencoder architectures} learn compact latent representations suitable for shape analysis and generation. AtlasNet's multiple parameterization approach covers complex surfaces, while FoldingNet achieves notable parameter efficiency (approximately 7\% of fully-connected alternatives). Both achieve strong transfer classification performance, indicating their representations capture meaningful shape semantics.

\textbf{Deformation-based methods} excel when template topology is known. They maintain mesh connectivity throughout deformation, avoiding self-intersections and enabling real-time performance in applications like robotic object tracking. The 58Hz tracking rate achieved by Real-NVP demonstrates practical viability for interactive systems.

\textbf{Primitive-based methods}, particularly BSP-Net, address a critical gap by generating compact, watertight meshes with sharp features. These methods are especially valuable for CAD applications where geometric precision is essential.

\subsection{Open Problems and Future Directions}

Several challenges remain for the research community:

\textbf{Topology Learning:} Current deformation methods cannot change topology, limiting applicability to shapes with known genus. Methods that can learn and modify connectivity during reconstruction would significantly expand the scope of learning-based approaches.

\textbf{Scene-level Reconstruction:} Most surveyed methods target single objects. Extending these approaches to handle complex scenes with multiple objects, occlusions, and varying scales remains challenging.

\textbf{Real-world Robustness:} While methods show strong performance on synthetic datasets, degradation on real sensor data (with noise, outliers, and non-uniform sampling) requires further investigation.

\textbf{Unified Evaluation:} The field lacks standardized benchmarks evaluating all aspects simultaneously---geometric accuracy, topological correctness, sharp feature preservation, and computational efficiency.

\textbf{Hybrid Approaches:} Combining strengths of different paradigms (e.g., primitive-based initialization with deformation-based refinement) may yield methods that are both geometrically accurate and efficient.

\subsection{Recommendations for Newcomers}

For practitioners entering this field, we recommend:
\begin{enumerate}
    \item Start with AtlasNet for general-purpose reconstruction---it offers a balance of quality, flexibility, and well-maintained code.
    \item Use PCN when dealing with incomplete data---its coarse-to-fine completion strategy handles partial inputs well.
    \item Consider BSP-Net for CAD applications where watertight outputs and sharp features are essential.
    \item Leverage the code repositories listed in Section~\ref{sec:resources} to accelerate implementation.
    \item Evaluate on both ShapeNet and ModelNet to assess generalization across synthetic datasets.
\end{enumerate}

As 3D sensing becomes increasingly prevalent in autonomous systems, robotics, and augmented reality, learning-based mesh reconstruction will continue to be an active area of research with significant practical impact.

\printbibliography

\end{document}